\documentclass{article}
\usepackage{ijcai22}

\usepackage{times}
\usepackage{soul}
\usepackage{url}
\usepackage[hidelinks]{hyperref}
\usepackage[utf8]{inputenc}
\usepackage[small]{caption}
\usepackage{graphicx}
\usepackage{amsmath}
\usepackage{amsthm}
\usepackage{booktabs}
\usepackage{algorithm}
\usepackage{algorithmic}
\usepackage{multirow}
\usepackage{bm}
\usepackage{amssymb}
\title{Eliciting Knowledge from Pretrained Language Models for Prototypical Prompt Verbalizer}

\author{
    Author Name
    \affiliations
    Affiliation
    \emails
    pcchair@ijcai-22.org
}
\author{
Yinyi Wei
\and
Tong Mo\footnote{Contact Author}
\and
Yongtao Jiang
\and
Weiping Li
\and
Wen Zhao
\affiliations
School of Software and Microelectronics, Peking University
\emails
wyyy@pku.edu.cn,
motong@ss.pku.edu.cn,
yongtao@pku.edu.cn,
wpli@ss.pku.edu.cn,
zhaowen@pku.edu.cn
}

\begin{document}

\maketitle

\begin{abstract}
  Recent advances on prompt-tuning cast few-shot classification tasks as a masked language modeling problem. By wrapping input into a template and using a verbalizer which constructs a mapping between label space and label word space, prompt-tuning can achieve excellent results in zero-shot and few-shot scenarios. However, typical prompt-tuning needs a manually designed verbalizer which requires domain expertise and human efforts. And the insufficient label space may introduce considerable bias into the results. In this paper, we focus on eliciting knowledge from pretrained language models and propose a prototypical prompt verbalizer for prompt-tuning. Labels are represented by prototypical embeddings in the feature space rather than by discrete words. The distances between the embedding at the masked position of input and prototypical embeddings are used as classification criterion. For zero-shot settings, knowledge is elicited from pretrained language models by a manually designed template to form initial prototypical embeddings. For few-shot settings, models are tuned to learn meaningful and interpretable prototypical embeddings. Our method optimizes models by contrastive learning. Extensive experimental results on several many-class text classification datasets with low-resource settings demonstrate the effectiveness of our approach compared with other verbalizer construction methods. Our implementation is available at \url{https://github.com/Ydongd/prototypical-prompt-verbalizer}.
 
\end{abstract}
\section{Introduction}
\begin{figure}[ht]
        \small
        \centering
        \includegraphics[width=1\linewidth]{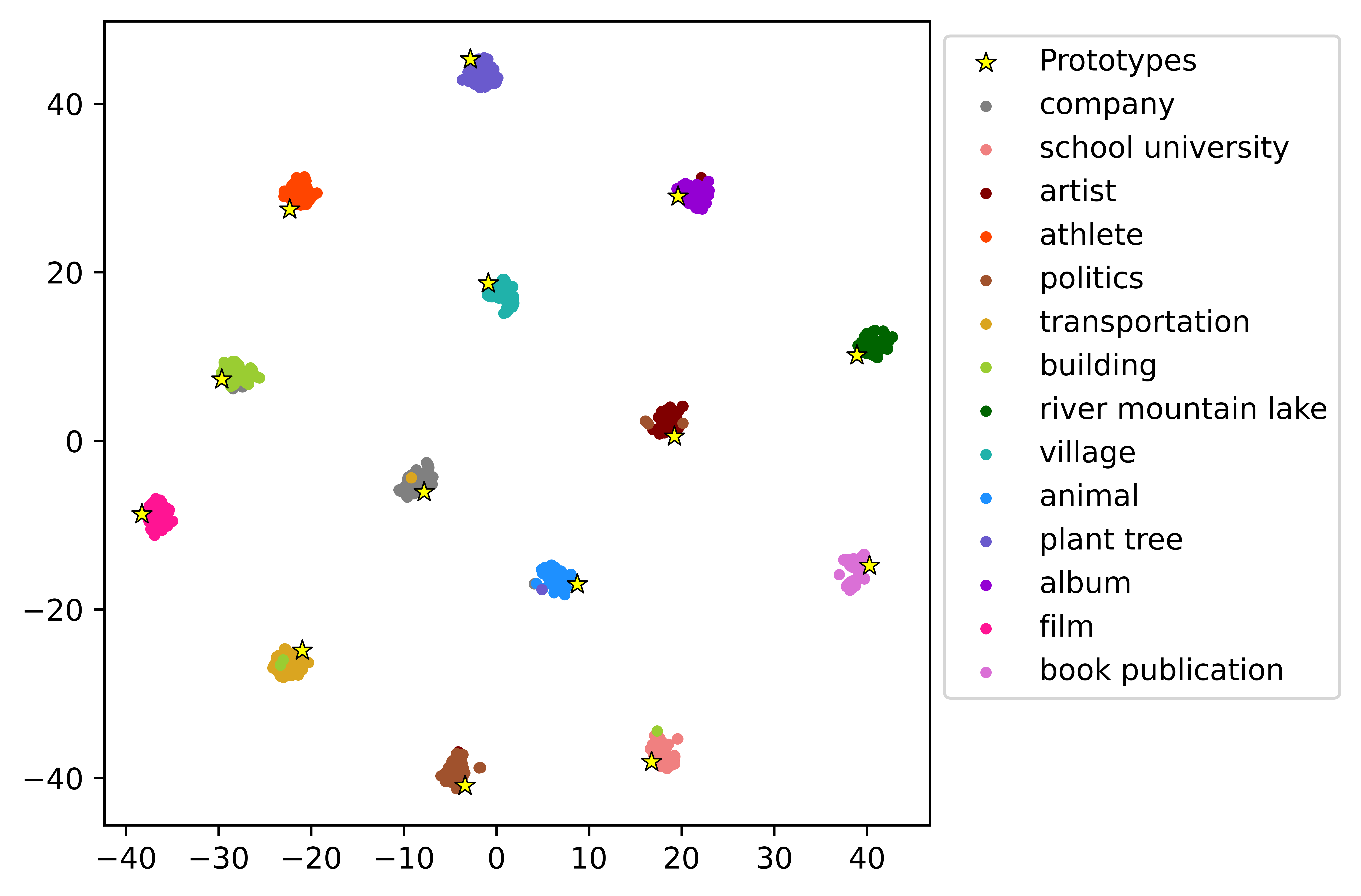}
        \caption{Prototypical embeddings and {\tt [MASK]} embeddings on DBPedia processed by prototypical prompt verbalizer and visualized after dimension reduction by $t$-SNE.}
        \label{fig2}
    \end{figure}

  In recent years, pretrained language models (PLMs) have shown excellent performance for language understanding and language generation in NLP tasks. By pretraining on large-scale corpora, models with rich semantics and knowledge can significantly benefit downstream tasks \cite{roberts2020much}. Due to the magnitude and potential of PLMs, it has become a topical issue how to motivate PLMs and appropriately elicit knowledge from them for downstream tasks.
  
  The most widely used method for downstream tasks is fine-tuning \cite{devlin2018bert}. By adding a classifier on the top of PLMs, fine-tuning has achieved remarkable results on supervised tasks compared with traditional methods. For an example, in the task of text classification, after taking the embedding of {\tt [CLS]} token and applying a classifier upon it, fine-tuning can easily obtain corresponding label for an input \cite{howard2018universal}. However, since the parameters of the classifier in fine-tuning are randomly initialized, it needs sufficient labeled data for training, thus fine-tuning is hard to obtain satisfactory results in scenarios with little labeled data and will hinder the transfer of knowledge in PLMs to downstream tasks.

  To alleviate this issue, prompt-tuning, a new paradigm for using PLMs, has been proposed for low-resource works to narrow the gap between pretraining tasks and downstream tasks \cite{schick2020exploiting}. The main idea of prompt-tuning is to transform a downstream task into a cloze question, which is consistent with the pretraining process of PLMs. Take text classification for an example, the input sentence is wrapped into a task-specific template, e.g.,``[Category: {\tt [MASK]}] {\tt [SENTENCE]}", where {\tt [SENTENCE]} is filled with the input sentence and {\tt [MASK]} is served as the set of predicted words. After constructing a verbalizer, a mapping between label space and label word space, the predicted words from {\tt [MASK]} can be easily transformed into corresponding labels. Since the verbalizer directly determines the effectiveness of classification, how to construct a verbalizer becomes a very important issue in prompt-tuning.
  
  The traditional verbalizer construction method uses a word corresponding to a label to construct an one-to-one mapping \cite{schick2020exploiting}, which requires domain expertise and human efforts to determine which word to choose to represent a label because incorrect mapping may lead to extremely significant bias. To mitigate the drawbacks of manually designed verbalizer, some works propose to search for the best label word for a label by gradient descent \cite{schick2020automatically}. But these approaches still construct an one-to-one mapping verbalizer, which can't summarize the semantics of a label well and makes the coverage of a label vulnerable. To expand the coverage of label words for a specific label, one-to-many mapping verbalizers which select related words from multiple large knowledge bases have been proposed \cite{hu2021knowledgeable}. Such an approach can greatly enhance the semantics of labels, but due to the excessive number of related words obtained from knowledge bases, related words of different labels may overlap, and deciding which related words are suitable for a certain task also requires domain expertise and human efforts.
  
  To eliminate the impact of discrete words, soft verbalizers have been proposed \cite{hambardzumyan2021warp,zhang2021differentiable}. Soft verbalizers treat labels as trainable tokens and the optimization objective is set to a cross entropy loss between output of masked language model and the label tokens. Label tokens can be considered as label embeddings to some extent. Such methods are difficult to form highly representational and meaningful label embeddings, and are also hard to apply in zero-shot and few-shot scenarios because both label embeddings and classifiers are randomly initialized, which is also known as a Cold Start problem.

  In this paper, we propose prototypical prompt verbalizer to address the above issues. By using prototypical networks, we generate prototypical embeddings for different labels in the feature space to summarize the semantic information of labels. With respect to the classification criterion, we compute distances between the embedding of the input's {\tt [MASK]} token and prototypical embeddings in the feature space, then select the label corresponding to embedding with the highest similarity as the label of input. To overcome the difficulties of application in scenarios with few labeled training samples when training from scratch, we use the word corresponding to each label in the instruction document and a small number of sentences containing this word in the unlabeled corpus to form initial prototypical embeddings by a manual template. Note that although we also use some of the label words here, we do not need to do operations such as filtering and expansion, we only use the label words to tackle the Cold Start problem, not to select the final label. Even though there is a lot of noise in the selected sentences containing specific words, the semantics of labels can still be extracted to some extent. Based on contrastive learning, we devise three different objective functions to optimize the models. Results on DBPedia after dimension reduction are shown in Figure \ref{fig2}. In summary, the main contributions of our work are:

\begin{itemize}
  \item We design a method which can generate prototypical embeddings for labels as semantic representations in the feature space and use contrastive learning at instance-instance and instance-label level to learn meaningful and interpretable prototypical embeddings.
  \item For zero-shot scenario, to tackle Cold Start problem, we use some unlabeled sentences containing specific words to generate initial prototypical embeddings.
  \item The results of extensive experiments on three many-class text classification datasets with low-resource settings demonstrate the effectiveness of our approach.
\end{itemize}

\begin{figure*}[t]
        \small
        \centering
        \includegraphics[width=0.85\linewidth]{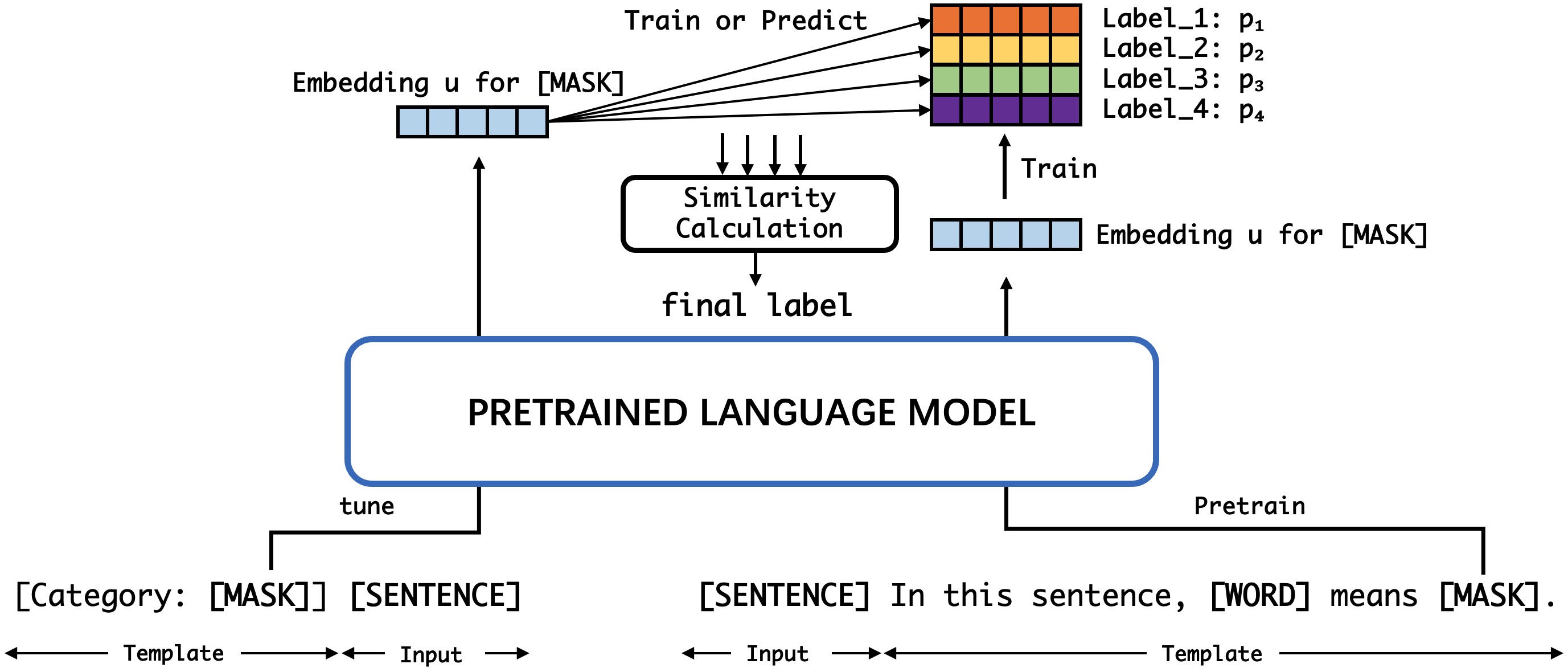}
        \caption{Overview of our method. The right side shows the pretraining process, where the model knowledge is elicited through a manual template combined with a specific word. The left side is the training process. Both pretraining and training process are trained with contrastive objective function.}
        \label{fig1}
    \end{figure*}

\section{Related Work}
\subsection{Prompt-tuning}
  Since there exists a huge gap between pretraining tasks and downstream tasks, some works introduce a new method named prompt-tuning to overcome it. GPT-3 \cite{brown2020language} shows that large-scale language models with prompt-tuning can achieve excellent performance in low-data environments. The work \cite{schick2020exploiting} shows that prompt-tuning can also perform superiorly in small-scale language models \cite{devlin2018bert,liu2019roberta}. While most of current works about prompt-tuning is in text classification tasks, some works have applied it to information extraction tasks \cite{cui2021template,han2021ptr,si2021generating} and text generation tasks \cite{li2021prefix}.

\subsection{Verbalizer Construction}
  In prompt-tuning, there are two key factors: template and verbalizer. When in low-resource settings, how to construct a good verbalizer becomes an essential factor in improving the efficiency of prompt-tuning. Current verbalizers are divided into two main categories: word-based verbalizers \cite{schick2020exploiting,schick2020automatically,hu2021knowledgeable} and embedding-based verbalziers \cite{hambardzumyan2021warp,zhang2021differentiable}. The former may lead to weak label coverage and insufficient semantics for labels. And the construction of such verbalizers may require domain expertise and human efforts. While the latter have a Cold Start problem and it is a challenging issue to form meaningful and interpretable label embeddings with current methods.

\subsection{Contrastive Learning}
  Contrastive learning \cite{hadsell2006dimensionality} aims to learn similar representations for positive instances and different representations for negative instances, and is widely used for self-supervised representation learning mainly in domain of computer vision \cite{wu2018unsupervised}. As for natural language processing, some well-known works also apply the idea of contrastive learning, such as Word2Vec \cite{mikolov2013efficient} and BERT \cite{devlin2018bert}. In NLP tasks, contrastive learning is usually used for generating high-quality text representations based on the construction of positive and negative samples \cite{gao2021simcse}.

\subsection{Prototypical Networks}
  In few-shot classification tasks, a classifier often needs to generate label representations with insufficient instances. To address this issue, some works \cite{snell2017prototypical,ji2020improved} have proposed to use prototypical networks to learn representations for labels in the feature space. In contrast to traditional approaches, prototypical networks can learn a metric space where classification can be performed by computing distances from the input representation to prototypical representations of labels. The approach introduces a semantic generalization of labels, which can achieve excellent results with limited data. Some works have applied prototypical networks to other domains, such as information extraction \cite{ding2021prototypical}.

\section{Method}
    In this section, we present our method to construct prototypical prompt verbalizer. A key motivation behind this is that, eliciting knowledge from pretrained language models and using it to generate prototypical embeddings for labels. Firstly, we describe general paradigm of prompt tuning. Secondly, We elaborate our prototypical prompt verbalizer in detail. Finally, we introduce the different settings in zero-shot and few-shot scenarios.

\subsection{Overview}
\subsubsection{General Prompt-tuning}
    Formally, denote $\mathcal{M}$, $\mathcal{T}$ and $\mathcal{V}$ as pretrained language model, template function and verbalizer function, respectively. Given an input $x$, template function $\mathcal{T}$ inserts pieces of texts into $x$ to convert it into the corresponding input of $\mathcal{M}$ which has a {\tt[MASK]} token in it, i.e., $x_{prompt} = \mathcal{T}(x)$. Let $V$ be the label words set, $Y$ be the label set. $\mathcal{V}:Y\rightarrow V$ is a mapping from label space to label word space, $\mathcal{V}(y)$ represents label words corresponding to label $y$. Then for input $x$, the probability of label $y$ is
    \begin{equation}
        P(y|x) = \sigma (P_\mathcal{M} ({\text{\tt [MASK]}} = v|v\in \mathcal{V}(y) ))
    \end{equation}
    where $\sigma(\cdot)$ determines which aggregation function to be used for labels with several different label words, such as $max$ or $average$.
    
    With prompt-tuning, a classification problem can be transferred into a masked language modeling problem by filling the {\tt[MASK]} token in the input.
    
    In order to include more semantic information for different labels, we propose prototypical prompt verbalizer based on contrastive learning to extend the scope of prompt-tuning.

\subsubsection{Prototypical Prompt Verbalizer}
    In prototypical prompt verbalizer, instead of directly predicting the corresponding label words from {\tt[MASK]} token, we first generate prototypical embeddings which capture the main semantic information of labels, then for each input, we compute similarity between the embedding of {\tt[MASK]} token and prototypical embeddings and finally select the label corresponding to the most similar prototypical embedding as the classification result.
    
    Given an input $x$, we first convert it into a template-based input with a {\tt[MASK]} token: $x_{prompt}=\mathcal{T}(x)$, then we feed $x_{prompt}$ into pretrained language model $\mathcal{M}$ and obtain the last layer's hidden state of output $\bm{h}=\mathcal{M}(x_{prompt})$. We take the embedding of the {\tt[MASK]} token $h_{\rm [MASK]} \in \mathbb{R}^M$ as the initial embedding for this input.
    
    For each initial embedding, in order to give it a more compact semantic representation, we use a transforming function $f:\mathbb{R}^M\rightarrow\mathbb{R}^D$ to map it to a new feature space:
    \begin{equation}
        f(h_{{\mbox{\tt [MASK]}}}) = u
    \end{equation}
    
    For each label $y\in Y$, we generate a prototypical embedding $p \in \mathbb{R}^D$ in the feature space to abstract the essential semantics of $y$. Transforming function and prototypical embeddings are trained from scratch.

    We use cosine similarity $d:\mathbb{R}^D\times\mathbb{R}^D\rightarrow [-1,1]$ to measure the similarity between transformed embeddings as $s(u_i, u_j)$ and between a transformed embedding and a prototypical embedding as $s(p, u)$:
    \begin{equation}
        s(u_i, u_j) = \frac{u_i\cdot u_j}{\left \|u_i\right \|\left \|u_j\right \|},\quad s(p, u) = \frac{p\cdot u}{\left \|p\right \|\left \|u\right \|}
    \end{equation}
    
    A batch is defined as $\mathcal{B}=\{u_0, u_1, \dots, u_n\}$, $u_i$ is the transformed embedding of the $i$-th input $x_i$, $|\mathcal{B}|=N$. In each batch, our intuition is to make individual embeddings and prototypical embeddings have meaningful and interpretable representations in the feature space. For this purpose, we define three contrastive objective functions.
    
    The first objective function aims to keep embeddings of the same kind close to each other and embeddings of different kinds away from each other. Inspired by \cite{soares2019matching} and \cite{ding2021prototypical}, we define it as:
    \begin{multline}
        \mathcal{L}_{s} =\\
        -\frac{1}{N^2}\sum_{i,j}{\rm log}\frac{\Theta(i,j)}{{\Theta(i,j) + \textstyle \sum_{j',\pi(j')\neq \pi(i)}}{\rm exp}(s(u_i,u_j'))} 
    \end{multline}
    where $\Theta(i,j)={\rm exp}(\varphi(i,j)s(u_i,u_j))$, $\varphi(i,j)$ is an indicator function to indicate whether $u_i$ and $u_j$ having the same label, i.e., given two embeddings $u_i$ and $u_j$, if these two embeddings have the same label, then $\varphi(i,j)=1$, otherwise 0. $\pi(i)$ denotes the label of $u_i$.
    
    The second and third objective funtions are used for learning prototypical embeddings of the labels. The second objective function makes an embedding close to the prototypical embedding to which it belongs and away from other prototypical embeddings. While the third objective function keeps a prototypical embedding in $\mathcal{B}$ away from other embeddings which have different labels from it.

    \begin{align}
        \mathcal{L}_{p_1} &= -\frac{1}{N}\sum_{i}{\rm log}\frac{{\rm exp}(u_i,p_{\pi(i)})}{ {\textstyle \sum_{k}{\rm exp}(u_i,p_k)} }  \\
        \mathcal{L}_{p_2} &= -\frac{1}{N}\sum_{i}{\rm log}\frac{{\rm exp}(p_{\pi(i)}, u_i)}{ {\textstyle \sum_{j,j=i|\pi(j')\neq \pi(i)}{\rm exp}(p_{\pi(i)},u_j)} }
    \end{align}
    
    Finally, combining the above three objective functions with hyperparameters $\lambda_1$, $\lambda_2$, $\lambda_3$, the full objective funtion is computed as:
    \begin{equation}
        \mathcal{L} = \lambda_1 \mathcal{L}_{s} + \lambda_2 \mathcal{L}_{p_1} + \lambda_3 \mathcal{L}_{p_2}
    \end{equation}
    
    Given an input $x$, $u$ is the corresponding embedding for $x$, then the probability for label $y$ is:
    \begin{equation}
        p(y|x) = \frac{{\rm exp}(s(u, p_y))}{\sum_{k}{\rm exp}(s(u, p_k))}
    \end{equation}
    
    Our method is shown in Figure \ref{fig1}, we will detail the use of our method in zero-shot and few-shot scenarios with pretraining and training process in following sections.
    
\subsection{zero-shot settings}
    In zero-shot learning settings, it is challenging to initialize prototypical embeddings for labels. Since pretrained language model $\mathcal{M}$ is trained on large-scale corpora and contains a lot of rich semantic information, we use a manually designed template to elicit knowledge from $\mathcal{M}$ to form initial prototypical embeddings for labels.
    
    For a specific label $y$, we use its corresponding literal word $v$ in the instruction document and sample a small amount of unlabeled sentences $\mathcal{Q} = \{q_1,q_2,\dots,q_Q\}$ containing the word $v$ from the training set with labels removed. Given a specific word $v$ and a sentence $q_i$, we wrap them into a template $\mathcal{T}_z$:``{\tt [SENTENCE]} In this sentence, {\tt [WORD]} means {\tt [MASK]}." where $q_i$ is for {\tt [SENTENCE]} blank and $v$ is for {\tt [WORD]} blank. 
    
    Then we take the embedding of {\tt [MASK]} token in the last layer's hidden state of $\mathcal{M}(\mathcal{T}_z(q_i))$ acting as initial prototypical embedding for $y$ to perform the optimizing process described in the previous section. In this way, the initial prototypical embeddings can be obtained and we name this process pretraining in our method. Randomly sampled sentences may be very noisy due to different meanings of a specific word, however, the probability of a specific word with different meanings appearing in one sampling process is relatively small, and to the purpose of simplicity, we do not prune them.

\subsection{few-shot settings}
    In few-shot learning settings, $\mathcal{T}_{\mathcal{D}}$ is the set of templates for dataset $\mathcal{D}$. For $\mathcal{T}_i \in \mathcal{T}_{\mathcal{D}}$, we simply wrap input into $\mathcal{T}_i$ and take the embedding of {\tt [MASK]} token in the last layer of $\mathcal{M}$'s output to form prototypical embeddings as mentioned above. We name this process training in our method.

\section{Experiments}
\begin{table}
    \small
    \centering
    \begin{tabular}{ccc}
    \toprule
    Name  & \# Class & Test Size \\
    \midrule
    AG's News   & 4  & 7600      \\
    Yahoo Answers & 10  & 60000      \\
    DBPedia     & 14  & 70000    \\
    \bottomrule
    \end{tabular}
    \caption{Statistics for AG's News, Yahoo Answers and DBPedia}
    \label{table1}
    \end{table}

\begin{table*}[ht]
    \small
    \centering
    \begin{tabular}{lllll}
    \toprule
    $k$                      & Method & AG's News & Yahoo Answers & DBPedia  \\
    \midrule
    \multirow{3}{*}{0 }    & PT & 71.84 {\scriptsize$\pm$ 5.82} (\textbf{80.36}) & 50.68 {\scriptsize$\pm$ 10.43} (59.90) & 65.10 {\scriptsize$\pm$ 4.43} (71.10) \\
                           & $\text {PPV}_{w/\;p}$(avg) & 67.12 {\scriptsize$\pm$ 9.07} (77.00) & 51.84 {\scriptsize$\pm$ 11.41} (\textbf{59.93}) & 78.86 {\scriptsize$\pm$ 3.41} (\textbf{83.14}) \\
                           & $\text {PPV}_{w/\;p}$(max) & \textbf{72.14} {\scriptsize$\pm$ 6.61} (77.00) & \textbf{53.85} {\scriptsize$\pm$ 7.91} (\textbf{59.93}) & \textbf{80.65} {\scriptsize$\pm$ 2.05} (\textbf{83.14}) \\
    \midrule
    \multirow{5}{*}{1    } & FT & 38.38 {\scriptsize$\pm$ 5.79} (45.23) & 16.50 {\scriptsize$\pm$ 3.09} (20.80) & 30.67 {\scriptsize$\pm$ 2.38} (33.56) \\
                           & PT & \textbf{77.69} {\scriptsize$\pm$ 8.16} (\textbf{85.72}) & 57.77 {\scriptsize$\pm$ 3.39} (\textbf{62.19}) & 93.97 {\scriptsize$\pm$ 1.18} (\textbf{95.96}) \\
                           & SPV & 35.82 {\scriptsize$\pm$ 6.82} (47.00) & 20.83 {\scriptsize$\pm$ 3.43} (25.68) & 64.77 {\scriptsize$\pm$ 11.17} (76.67) \\
                           & $\text {PPV}_{w/\;p}$ & 74.29 {\scriptsize$\pm$ 5.52} (80.27) & \textbf{57.79} {\scriptsize$\pm$ 1.54} (60.16) & \textbf{94.21} {\scriptsize$\pm$ 0.50} (94.96) \\
                           & $\text {PPV}_{w/o\;p}$ & 57.10 {\scriptsize$\pm$ 6.34} (70.85) & 24.21 {\scriptsize$\pm$ 3.23} (28.84) & 61.06 {\scriptsize$\pm$ 4.47} (70.99) \\
    \midrule
    \multirow{5}{*}{5    } & FT & 62.56 {\scriptsize$\pm$ 16.02} (75.03) & 56.09 {\scriptsize$\pm$ 0.41} (56.65) & 94.48 {\scriptsize$\pm$ 0.87} (95.68) \\
                           & PT & \textbf{83.76} {\scriptsize$\pm$ 2.08} (\textbf{86.88}) & 61.61 {\scriptsize$\pm$ 1.94} (\textbf{65.70}) & 95.90 {\scriptsize$\pm$ 0.77} (96.85) \\
                           & SPV & 57.81 {\scriptsize$\pm$ 7.51} (68.96) & 46.67 {\scriptsize$\pm$ 8.37} (57.61) & 94.54 {\scriptsize$\pm$ 2.01} (\textbf{97.49}) \\
                           & $\text {PPV}_{w/\;p}$ & 81.49 {\scriptsize$\pm$ 1.59} (83.52) & \textbf{63.06} {\scriptsize$\pm$ 1.55} (65.28) & \textbf{96.54} {\scriptsize$\pm$ 0.41} (97.22) \\
                           & $\text {PPV}_{w/o\;p}$ & 79.00 {\scriptsize$\pm$ 5.13} (84.03) & 56.95 {\scriptsize$\pm$ 5.45} (63.77) & 95.46 {\scriptsize$\pm$ 1.00} (96.95) \\
    \midrule
    \multirow{5}{*}{10}    & FT & 82.76 {\scriptsize$\pm$ 0.35} (83.07) & 62.57 {\scriptsize$\pm$ 1.16} (63.75) & 97.61 {\scriptsize$\pm$ 0.41} (98.05) \\
                           & PT & 84.27 {\scriptsize$\pm$ 2.02} (\textbf{87.30}) & 64.19 {\scriptsize$\pm$ 1.46} (65.94) & 97.06 {\scriptsize$\pm$ 0.80} (98.10) \\
                           & SPV & 73.08 {\scriptsize$\pm$ 5.25} (80.48) & 59.68 {\scriptsize$\pm$ 2.29} (62.86) & 97.22 {\scriptsize$\pm$ 0.38} (97.87) \\
                           & $\text {PPV}_{w/\;p}$ & 84.39 {\scriptsize$\pm$ 1.11} (86.25) & 65.05 {\scriptsize$\pm$ 1.36} (\textbf{67.40}) & 97.76 {\scriptsize$\pm$ 0.25} (98.32) \\
                           & $\text {PPV}_{w/o\;p}$ & \textbf{84.73} {\scriptsize$\pm$ 1.15} (86.28) & \textbf{65.41} {\scriptsize$\pm$ 0.71} (66.39) & \textbf{97.89} {\scriptsize$\pm$ 0.29} (\textbf{98.34}) \\
    \midrule
    \multirow{5}{*}{20}    & FT & 85.23 {\scriptsize$\pm$ 0.18} (85.44) & 66.85 {\scriptsize$\pm$ 0.34} (67.29) & 98.01 {\scriptsize$\pm$ 0.19} (98.15) \\
                           & PT & 86.23 {\scriptsize$\pm$ 1.28} (88.03) & 67.11 {\scriptsize$\pm$ 0.66} (68.61) & 98.11 {\scriptsize$\pm$ 0.18} (98.32) \\
                           & SPV & 82.52 {\scriptsize$\pm$ 2.69} (85.48) & 65.78 {\scriptsize$\pm$ 1.28} (68.00) & 98.01 {\scriptsize$\pm$ 0.26} (98.44) \\
                           & $\text {PPV}_{w/\;p}$ & 86.84 {\scriptsize$\pm$ 0.92} (\textbf{88.40}) & 67.80 {\scriptsize$\pm$ 0.73} (69.00) & 98.25 {\scriptsize$\pm$ 0.23} (98.48) \\
                           & $\text {PPV}_{w/o\;p}$ & \textbf{86.92} {\scriptsize$\pm$ 0.79} (88.11) & \textbf{68.22} {\scriptsize$\pm$ 0.91} (\textbf{69.87}) & \textbf{98.34} {\scriptsize$\pm$ 0.19} (\textbf{98.63}) \\
    \bottomrule
    \end{tabular}
    \caption{Micro-F1 and standard deviation on AG's News, Yahoo Answers and DBPedia in zero and few-shot scenarios. The best Micro-F1 scores are shown in the brackets. The best results among all methods for the same $k$-shot experiment are marked in bold. FT represents fine-tuning. PT represents prompt-tuning. SPV represents soft prompt verbalizer. PPV represents our prototypical prompt verbalizer. $w/\; p$ and $w/o\; p$ are whether to apply pretraining process for prototypical prompt verbalizer respecively. We conduct experiments on three different random seeds for four different templates. In zero-shot scenario, the average results of all random seeds and the best results of one random seed are shown as avg and max respectively.}
    \label{table2}
    \end{table*}
    
    In this section, we conduct experiments on three many-class text classification datasets to empirically demonstrate the effectiveness of our prototypical prompt verbalizer.
\subsection{datasets}
    We evaluate our proposed method on three widely-used topic classification datasets: AG's News, Yahoo Answers \cite{zhang2015character} and DBPedia \cite{lehmann2015dbpedia}. Statistics of these datasets are shown in Tabel \ref{table1}.
    
    Due to the rich semantics and the high adaptability to different datasets, manual templates have an advantage over automatically generated templates in zero-shot and few-shot scenarios. To alleviate the bias in the results caused by different templates, we use four manual templates for each datasets as in \cite{schick2020exploiting} and \cite{hu2021knowledgeable}.

\subsection{Baselines}
    In this subsection, we introduce the baselines we use to demonstrate the effectiveness of our approach, including fine-tuning, general prompt-tuning and soft prompt verbalizer. We compare the baselines with our prototypical prompt verbalizer in both pretraining and without pretraining cases.
    
    {\bf Fine-tuning}. As the most popular paradigm for using pretrained language models, fine-tuning feeds the embedding in the last layer's hidden state of {\tt [CLS]} token to a classifier to obtain the final label of input. We do not conduct zero-shot tests on fine-tuning, since the parameters of its classifier are randomly initialized.
    
    {\bf Prompt-tuning}. As mentioned previously, prompt-tuning, a new paradigm that has emerged recently, can work well with little training data with the help of pretrained masked language head. For each label, we use the words from the instruction document as its label words and test it in zero-shot and few-shot settings. In our implementation, there may be multiple label words corresponding to one label, so the verbalizer here is not a simple one-to-one mapping but one-to-many mapping for some labels.
    
    {\bf Soft Prompt Verbalizer}. Soft prompt verbalizer treats labels as trainable tokens to mitigate the impact of discrete words. In our implementation, soft prompt verbalizer feeds the embedding of {\tt [MASK]} token to a classifier to obtain the final label. The optimization objective is cross entropy loss. Since the parameters of the classifier are randomly initialized, the approach is also not suitable for zero-shot scenario.

\subsection{Implementation Details}
    In zero-shot scenario, models evaluate on the entire test set without training on the labeled training data and we cut sentences from unlabeled corpus by nltk \cite{bird2009natural}. While in few-shot scenario, we carry out 1, 5, 10 and 20-shot experiments. For a $k$-shot experiment, We randomly select $k$ instances of each class from the training set as the new training set and test the model on the entire test set.
    
    When pretraining prototypical prompt verbalizer, we sample 60 sentences for AG's News, 40 sentences for Yahoo Answers, 30 sentences for DBPedia.
    
    For all experiments, we use RoBERTa large \cite{liu2019roberta} as pretrained language model and use Micro-F1 as test metrics. For fine-tuning, prompt-tuning and prototypical prompt verbalizer, we use our own framework. For soft prompt verbalizer, we use OpenPrompt framework \cite{ding2021openprompt}. We select AdamW with the learning rate of $3e-5$ for optimazion. The size of transformed embedding is set to 256 and the max sequence length is set to 512. We train the model for 10 epochs with the batchsize setting to 8 in each experiment.

\subsection{Results and analysis}
\subsubsection{Experimental results}
    In this subsection, We detail the results of our method and perform an insightful analysis. Experimental results are shown in Table \ref{table2}.
    
    In zero-shot scenario, since the randomly sampled sentences used for pretraining are noisy, which leads to unexpected deviations in the semantics of the generated initial prototypical embeddings, the results of pretraining on various sentences differ to some extent. However, the results still indicate effectiveness. On average, our method works better on DBPedia and Yahoo Answers compared with prompt-tuning. In terms of the best results, our method outperforms prompt-tuning on all three datasets. The results show that the pretraining process can elicit knowledge from pretrained language model well, and to some extent matches the masked language model head trained on large-scale corpora.
    
    In few-shot scenario, our method is proven to be powerful. Prototypical prompt verbalizer without pretraining beats fine-tuning and soft prompt verbalizer on 1-shot and 5-shot experiments, but the results are weaker than prompt-tuning due to insufficient samples for contrastive learning. Comparison with soft prompt verbalizer demonstrates that our method can generate high quality, meaningful and interpretable prototypical embeddings in the feature space. With pretraining process, prototypical prompt verbalizer can obtain initial prototypical embeddings for labels and achieve better results in comparison to prompt tuning except on AG's News. We attribute the reason why our method do not work as well as prompt-tuning in 1-shot and 5-shot scenarios on AG's News to too few categories and training samples which result in inadequate contrastive learning. When in 10-shot and 20-shot scenarios, our method outperforms prompt tuning with or without pretraining, however, unpretrained one can attain better results with respect to the pretrained one. The major reason is that the semantics of initial prototypical embeddings generated in the pretraining process are a bit different from the real prototypical embeddings, which takes more time to reach the global optimum and also has a higher probability of falling into the local optimum.

\subsubsection{Freezing the pretrained language model}
    \begin{table}[t]
    \small
    \centering
    \begin{tabular}{lllc}
    \toprule
    method          & 5-shot & 20-shot & \# param ($\times 10^4$) \\
    \midrule
    $\text{SPV}$            & 18.66 {\scriptsize$\pm$ 2.88} & 37.06 {\scriptsize$\pm$ 4.11} & 1.024 \\
    $\text{PT}$             & 58.49 {\scriptsize$\pm$ 5.42} & 62.82 {\scriptsize$\pm$ 3.88} & 110.2(5147.1)  \\
    $\text{PPV}_{w/o\;p}$   & 11.91 {\scriptsize$\pm$ 2.37} & 17.74 {\scriptsize$\pm$ 6.39} & 26.4 \\
    $\text{PPV}_{w/\;p}$    & 59.13 {\scriptsize$\pm$ 1.94} & 61.02 {\scriptsize$\pm$ 1.99} & 26.4 \\
    \bottomrule
    \end{tabular}
    \caption{Results on Yahoo Answers with trainable parameters after freezing the pretrained language model. Since the mask language head of prompt tuning has parameters tied to the input layer which do not participate in the optimization process, we list it in the brackets.}
    \label{table3}
    \end{table}

    We further freeze the pretrained language model and conduct experiments on Yahoo Answers as illustrated in Table \ref{table3} with the number of head parameters. The goal of prototypical prompt verbalizer is to form prototypical embeddings and construct a mapping between embeddings output from pretrained model and the feature space, where limited trainable parameters will make it hard to train. Soft prompt verbalizer, on the other hand, only needs to construct a mapping from the output embeddings to the label space and is therefore easier to train. After pretraining, prototypical prompt verbalizer basically forms prototypical embeddings and the mapping, so it obtains fine results. However, due to the numerous parameters of mask language head, prompt-tuning will achieve better results when the size of training set increases.

\subsubsection{Effect of objective function}
    \begin{table}[t]
    \small
    \centering
    \begin{tabular}{llll}
    \toprule
    loss  & 0-shot & 5-shot & 20-shot  \\
    \midrule
    $\mathcal{L}_s$                                     & 11.62 {\scriptsize$\pm$ 2.35} & 9.12 {\scriptsize$\pm$ 2.64} & 6.27 {\scriptsize$\pm$ 0.85} \\
    $\mathcal{L}_{p_1} + \mathcal{L}_{p_2}$                     & 39.63 {\scriptsize$\pm$ 17.38} & 59.51 {\scriptsize$\pm$ 2.86} & 67.88 {\scriptsize$\pm$ 0.46} \\
    $\mathcal{L}_s + \mathcal{L}_{p_1}$                     & 45.77 {\scriptsize$\pm$ 17.11} & 59.58 {\scriptsize$\pm$ 2.42} & 68.32 {\scriptsize$\pm$ 0.45} \\
    $\mathcal{L}_s + \mathcal{L}_{p_2}$                  & 41.89 {\scriptsize$\pm$ 17.35} & 58.02 {\scriptsize$\pm$ 2.48} & 68.27 {\scriptsize$\pm$ 0.41} \\   
    $\mathcal{L}_s + \mathcal{L}_{p_1} + \mathcal{L}_{p_2}$     & 48.64 {\scriptsize$\pm$ 16.72} & 60.70 {\scriptsize$\pm$ 2.31} & 68.55 {\scriptsize$\pm$ 0.45} \\
    \bottomrule
    \end{tabular}
    \caption{Results after fixing a random seed on Yahoo Answers with different combination of losses.}
    \label{table4}
    \end{table}

    To illustrate the effectiveness of our objective function, we conduct experiments with different combination of losses. As shown in Table \ref{table4}, $\mathcal{L}_{p_1}$ and $\mathcal{L}_{p_2}$ are used to form prototypical embeddings while $\mathcal{L}_s$ allows embeddings with identical labels to be aggregataed and embeddings with different labels to be dispersed in the feature space. 
    
\subsubsection{Pretraining on other data sources}
    \begin{table}[t]
    \small
    \centering
    \begin{tabular}{llll}
    \toprule
    source     & AG's News & Yahoo Answers & DBPedia  \\
    \midrule
    Unlabeled Set       & 67.12 {\scriptsize$\pm$ 9.07} & 51.84 {\scriptsize$\pm$ 11.41} & 78.86 {\scriptsize$\pm$ 3.41} \\
    Wikidata   & 64.19 {\scriptsize$\pm$ 6.67} & 49.95 {\scriptsize$\pm$ 6.21} & 76.57 {\scriptsize$\pm$ 7.18} \\
    \bottomrule
    \end{tabular}
    \caption{Results on three datasets in zero-shot scenario with pretraining on different data sources.}
    \label{table5}
    \end{table}
    
    To explore whether the pretraining process works on other data sources as well, we conduct experiments with unlabeled training set and a small part of Wikidata as shown in Table \ref{table5}, and it is notable that DBPedia is also derived from Wikidata. The results illustrate that pretraining is also valid on other data sources. And no matter what data source is used, fluctuations can be significant due to the large noise in the pretraining process.

\subsection{Conclusion}
    In this paper, we propose prototypical prompt verbalizer to enhance the semantic scope of labels by forming prototypical embeddings and construct a mapping from output of pretrained language models to the feature space. To obtain meaningful and interpretable embeddings, we optimize models with contrastive objective functions. In order to solve the problem of poor results caused by parameter initialization in zero-shot and some few-shot scenarios, we propose to conduct pretraining on a small amount of unlabeled training set. The experiments show the effectiveness and potential of our method. However, the existence of large noise in randomly sampled sentences may seriously affect the pretraining results, and we will mitigate this issue in the future with denoising or self-supervision measures.

\newpage

\bibliographystyle{named}
\bibliography{ijcai22}

\begin{thebibliography}{}

\bibitem[\protect\citeauthoryear{Bird \bgroup \em et al.\egroup
  }{2009}]{bird2009natural}
Steven Bird, Ewan Klein, and Edward Loper.
\newblock {\em Natural language processing with Python: analyzing text with the
  natural language toolkit}.
\newblock " O'Reilly Media, Inc.", 2009.

\bibitem[\protect\citeauthoryear{Brown \bgroup \em et al.\egroup
  }{2020}]{brown2020language}
Tom~B Brown, Benjamin Mann, Nick Ryder, Melanie Subbiah, Jared Kaplan, Prafulla
  Dhariwal, Arvind Neelakantan, Pranav Shyam, Girish Sastry, Amanda Askell,
  et~al.
\newblock Language models are few-shot learners.
\newblock {\em arXiv preprint arXiv:2005.14165}, 2020.

\bibitem[\protect\citeauthoryear{Cui \bgroup \em et al.\egroup
  }{2021}]{cui2021template}
Leyang Cui, Yu~Wu, Jian Liu, Sen Yang, and Yue Zhang.
\newblock Template-based named entity recognition using bart.
\newblock {\em arXiv preprint arXiv:2106.01760}, 2021.

\bibitem[\protect\citeauthoryear{Devlin \bgroup \em et al.\egroup
  }{2018}]{devlin2018bert}
Jacob Devlin, Ming-Wei Chang, Kenton Lee, and Kristina Toutanova.
\newblock Bert: Pre-training of deep bidirectional transformers for language
  understanding.
\newblock {\em arXiv preprint arXiv:1810.04805}, 2018.

\bibitem[\protect\citeauthoryear{Ding \bgroup \em et al.\egroup
  }{2021a}]{ding2021openprompt}
Ning Ding, Shengding Hu, Weilin Zhao, Yulin Chen, Zhiyuan Liu, Hai-Tao Zheng,
  and Maosong Sun.
\newblock Openprompt: An open-source framework for prompt-learning.
\newblock {\em arXiv preprint arXiv:2111.01998}, 2021.

\bibitem[\protect\citeauthoryear{Ding \bgroup \em et al.\egroup
  }{2021b}]{ding2021prototypical}
Ning Ding, Xiaobin Wang, Yao Fu, Guangwei Xu, Rui Wang, Pengjun Xie, Ying Shen,
  Fei Huang, Hai-Tao Zheng, and Rui Zhang.
\newblock Prototypical representation learning for relation extraction.
\newblock {\em arXiv preprint arXiv:2103.11647}, 2021.

\bibitem[\protect\citeauthoryear{Gao \bgroup \em et al.\egroup
  }{2021}]{gao2021simcse}
Tianyu Gao, Xingcheng Yao, and Danqi Chen.
\newblock Simcse: Simple contrastive learning of sentence embeddings.
\newblock {\em arXiv preprint arXiv:2104.08821}, 2021.

\bibitem[\protect\citeauthoryear{Hadsell \bgroup \em et al.\egroup
  }{2006}]{hadsell2006dimensionality}
Raia Hadsell, Sumit Chopra, and Yann LeCun.
\newblock Dimensionality reduction by learning an invariant mapping.
\newblock In {\em 2006 IEEE Computer Society Conference on Computer Vision and
  Pattern Recognition (CVPR'06)}, volume~2, pages 1735--1742. IEEE, 2006.

\bibitem[\protect\citeauthoryear{Hambardzumyan \bgroup \em et al.\egroup
  }{2021}]{hambardzumyan2021warp}
Karen Hambardzumyan, Hrant Khachatrian, and Jonathan May.
\newblock Warp: Word-level adversarial reprogramming.
\newblock {\em arXiv preprint arXiv:2101.00121}, 2021.

\bibitem[\protect\citeauthoryear{Han \bgroup \em et al.\egroup
  }{2021}]{han2021ptr}
Xu~Han, Weilin Zhao, Ning Ding, Zhiyuan Liu, and Maosong Sun.
\newblock Ptr: Prompt tuning with rules for text classification.
\newblock {\em arXiv preprint arXiv:2105.11259}, 2021.

\bibitem[\protect\citeauthoryear{Howard and Ruder}{2018}]{howard2018universal}
Jeremy Howard and Sebastian Ruder.
\newblock Universal language model fine-tuning for text classification.
\newblock {\em arXiv preprint arXiv:1801.06146}, 2018.

\bibitem[\protect\citeauthoryear{Hu \bgroup \em et al.\egroup
  }{2021}]{hu2021knowledgeable}
Shengding Hu, Ning Ding, Huadong Wang, Zhiyuan Liu, Juanzi Li, and Maosong Sun.
\newblock Knowledgeable prompt-tuning: Incorporating knowledge into prompt
  verbalizer for text classification.
\newblock {\em arXiv preprint arXiv:2108.02035}, 2021.

\bibitem[\protect\citeauthoryear{Ji \bgroup \em et al.\egroup
  }{2020}]{ji2020improved}
Zhong Ji, Xingliang Chai, Yunlong Yu, Yanwei Pang, and Zhongfei Zhang.
\newblock Improved prototypical networks for few-shot learning.
\newblock {\em Pattern Recognition Letters}, 140:81--87, 2020.

\bibitem[\protect\citeauthoryear{Lehmann \bgroup \em et al.\egroup
  }{2015}]{lehmann2015dbpedia}
Jens Lehmann, Robert Isele, Max Jakob, Anja Jentzsch, Dimitris Kontokostas,
  Pablo~N Mendes, Sebastian Hellmann, Mohamed Morsey, Patrick Van~Kleef,
  S{\"o}ren Auer, et~al.
\newblock Dbpedia--a large-scale, multilingual knowledge base extracted from
  wikipedia.
\newblock {\em Semantic web}, 6(2):167--195, 2015.

\bibitem[\protect\citeauthoryear{Li and Liang}{2021}]{li2021prefix}
Xiang~Lisa Li and Percy Liang.
\newblock Prefix-tuning: Optimizing continuous prompts for generation.
\newblock {\em arXiv preprint arXiv:2101.00190}, 2021.

\bibitem[\protect\citeauthoryear{Liu \bgroup \em et al.\egroup
  }{2019}]{liu2019roberta}
Yinhan Liu, Myle Ott, Naman Goyal, Jingfei Du, Mandar Joshi, Danqi Chen, Omer
  Levy, Mike Lewis, Luke Zettlemoyer, and Veselin Stoyanov.
\newblock Roberta: A robustly optimized bert pretraining approach.
\newblock {\em arXiv preprint arXiv:1907.11692}, 2019.

\bibitem[\protect\citeauthoryear{Mikolov \bgroup \em et al.\egroup
  }{2013}]{mikolov2013efficient}
Tomas Mikolov, Kai Chen, Greg Corrado, and Jeffrey Dean.
\newblock Efficient estimation of word representations in vector space.
\newblock {\em arXiv preprint arXiv:1301.3781}, 2013.

\bibitem[\protect\citeauthoryear{Roberts \bgroup \em et al.\egroup
  }{2020}]{roberts2020much}
Adam Roberts, Colin Raffel, and Noam Shazeer.
\newblock How much knowledge can you pack into the parameters of a language
  model?
\newblock {\em arXiv preprint arXiv:2002.08910}, 2020.

\bibitem[\protect\citeauthoryear{Schick and
  Sch{\"u}tze}{2020}]{schick2020exploiting}
Timo Schick and Hinrich Sch{\"u}tze.
\newblock Exploiting cloze questions for few shot text classification and
  natural language inference.
\newblock {\em arXiv preprint arXiv:2001.07676}, 2020.

\bibitem[\protect\citeauthoryear{Schick \bgroup \em et al.\egroup
  }{2020}]{schick2020automatically}
Timo Schick, Helmut Schmid, and Hinrich Sch{\"u}tze.
\newblock Automatically identifying words that can serve as labels for few-shot
  text classification.
\newblock {\em arXiv preprint arXiv:2010.13641}, 2020.

\bibitem[\protect\citeauthoryear{Si \bgroup \em et al.\egroup
  }{2021}]{si2021generating}
Jinghui Si, Xutan Peng, Chen Li, Haotian Xu, and Jianxin Li.
\newblock Generating disentangled arguments with prompts: A simple event
  extraction framework that works.
\newblock {\em arXiv preprint arXiv:2110.04525}, 2021.

\bibitem[\protect\citeauthoryear{Snell \bgroup \em et al.\egroup
  }{2017}]{snell2017prototypical}
Jake Snell, Kevin Swersky, and Richard~S Zemel.
\newblock Prototypical networks for few-shot learning.
\newblock {\em arXiv preprint arXiv:1703.05175}, 2017.

\bibitem[\protect\citeauthoryear{Soares \bgroup \em et al.\egroup
  }{2019}]{soares2019matching}
Livio~Baldini Soares, Nicholas FitzGerald, Jeffrey Ling, and Tom Kwiatkowski.
\newblock Matching the blanks: Distributional similarity for relation learning.
\newblock {\em arXiv preprint arXiv:1906.03158}, 2019.

\bibitem[\protect\citeauthoryear{Wu \bgroup \em et al.\egroup
  }{2018}]{wu2018unsupervised}
Zhirong Wu, Yuanjun Xiong, Stella~X Yu, and Dahua Lin.
\newblock Unsupervised feature learning via non-parametric instance
  discrimination.
\newblock In {\em Proceedings of the IEEE conference on computer vision and
  pattern recognition}, pages 3733--3742, 2018.

\bibitem[\protect\citeauthoryear{Zhang \bgroup \em et al.\egroup
  }{2015}]{zhang2015character}
Xiang Zhang, Junbo Zhao, and Yann LeCun.
\newblock Character-level convolutional networks for text classification.
\newblock {\em Advances in neural information processing systems}, 28:649--657,
  2015.

\bibitem[\protect\citeauthoryear{Zhang \bgroup \em et al.\egroup
  }{2021}]{zhang2021differentiable}
Ningyu Zhang, Luoqiu Li, Xiang Chen, Shumin Deng, Zhen Bi, Chuanqi Tan, Fei
  Huang, and Huajun Chen.
\newblock Differentiable prompt makes pre-trained language models better
  few-shot learners.
\newblock {\em arXiv preprint arXiv:2108.13161}, 2021.

\end{thebibliography}

\end{document}